# An Automatic Method for Complete Brain Matter Segmentation from Multislice CT scan

Soumi Ray[a], Vinod Kumar[a], Chirag Ahuja[b], Niranjan Khandelwal[b]

[a]Biomedical Engineering Lab, Department of Electrical Engineering, Indian Institute of Technology Roorkee, 247667, Roorkee, Uttrakhand, India
[b]Department of Radiodiagnosis, Post graduate Institute of Medical Education and Research, Chandigarh, India

Abstract:

Importance of computed tomography (CT) images lies in imaging speed, image contrast & resolution and cost. Thus it has wide use in detection and diagnosis of brain diseases. But unfortunately reported works on CT segmentation is not very significant. In this paper, a robust automatic segmentation system is presented which is capable to segment complete brain matter from CT slices, without any lose in information. Proposed method is simple, fast, accurate and completely automatic. It can handle multislice CT scan in single run. From a given multislice CT dataset, one slice is selected automatically to form masks for segmentation. Two types of masks are created to handle nasal slices in a better way. Masks are created from selected reference slice using automatic seed point selection and region growing technique. One mask is designed for brain matter and another includes the skull of the reference slice. This second mask is used as global reference mask for all slices whereas the brain matter mask is implemented on only adjacent slices and continuously modified for better segmentation. Slices in given dataset are divided into two batches – before reference slice and after reference slice. Each batch segmented separately. Successive propagation of brain matter mask has demonstrated very high potential in reported segmentation. Presented result shows highest sensitivity and more than 96% accuracy in all cases. Resulted segmented images can be used for any brain disease diagnosis or further image analysis.

Keywords:  automatic segmentation; lossless segmentation; successive masking; CT images;

## 1. Introduction

Among all worldwide uninterrupted main stream researches research in biomedical medical instrumentation is under prime importance for its significance in betterment of mankind. Medical helps can be divided into two basic parts – diagnosis and treatment. For diagnosis several instruments are designed starting from stethoscope, medical weighing machine, blood pressure measuring kit, ECG (ElectroCardioGram) electrodes etc. to sophisticated X-ray, CT, MRI etc. Working principles of basic instruments like stethoscope, blood pressure measuring kit, ECG kit depends on signals acquired from different body parts, whereas the latter are designed using imaging technology. Internal images of human body are taken for observation, inspection and understanding for diagnosis purpose. Image contains more details and it is easy to understand. That is why, now-a-days, for complex or serious problems doctors depends on scanning i.e. imaging to avoid any misunderstanding of symptoms and wrong diagnosis.

      When any body part is scanned, several layers are imaged one after another on a 2D plane. Special training is required to understand such images properly. When brain, the only spherical part of body, is imaged the result turns into a very complex one. To get understandable images of complete brain, several cross-sectional images are taken one after another at particular predefined gap. Even after taking several images the complexity is not nullified and facial parts also get scanned in few images and change the pattern of information in an image. As brain contains soft tissues, MRI and CT scan instead of traditional x-ray offer very good results. In this paper reported work is limited to CT brain images only; but the concepts discussed here are also applicable for MRI. The reasons of selecting CT over any other modality are its low cost (Rehana et al., 2013), fast scanning procedure and wider availability in hospitals even if in rural areas. CT also has advantages over MRI for patients with claustrophobia and/or metal implants in body and less sensitive to movements (Al-ayyoub at al., 2013). This painless imaging offers information about soft tissue, blood vessels, cerebrospinal fluid (CSF) as well as bone at the same time with clear distinguishable boundaries. CT images with sufficiently accurate detail information about soft tissues support in depth investigation.

Clinical 2D images are further converted into a mathematical description for computational purpose. Any kind of digital image is actually a 2D numeric array. This array contains intensity information about each smallest bit i.e. pixel of all possible locations. The array size depends on the size of acquired image and its resolution. The image and hence the array can be square or rectangle. For any other shape, left out parts are considered as background. For an image of nXm size, the resolution is n*m. For mathematical analysis of an image, several array operations are performed. Any image array contains two significant types of data – background and information. For medical analysis the information part is important; but background also supports the computation process indirectly. Scan images often contain some unwanted information which can introduce errors during computation and analysis through computer. So, images are pre-processed to reduce error in final analysis.

In brain CT scan images, other than the brain matter, skull, soft tissue edema, parts of skin outside the skull and headrest also get scanned and introduce visible information. Advance CT machines store images in Digital Imaging and Communication in Medicine (DICOM) format which is an international standard for medical images (Varma, 2012). When images saved from DICOM to other format, sometimes patient's information gets printed on the background area of the scanned image. This information is undesirable and can introduce error in computerized analysis. Such unwanted parts in the scan must be removed before processing the image for analysis and diagnosis. Removal can be done by appropriate segmentation of the required parts from head CT scan.

## 2. State of Art

As discussed above, in a CT scan soft tissue, blood vessels, cerebrospinal fluid, skull, soft tissue edema if any, parts of skin outside the skull, air and most of the time headrest also get included and introduce some particular range of intensity. Skull and headrest shows highest intensity whereas CSF and air have lowest. Blood vessels, soft tissue, edema, skin etc. has overlapping intensity and so not distinguishable easily. For accurate, errorless computation through computer the removal of artifacts having similar intensity but no actual contribution is thus very important. Higher intensity skull and headrest is also removed to reduce computational complexity by reducing weight of image and to increase efficiency.

Though brain CT images are very important clinically, number of reported works on CT segmentation is not very large (Shahangian and Pourghassem, 2015). Moreover those reported segmentation works are majorly focused on any particular clinical abnormality like hemorrhage (Chen et al. 2009), tumor (Bardera et al. 2009), or have proposed segmentation of different parts of brain(Huang and Parra, 2015). Segmentation of tumor is mainly done on MR images(Anand and Kaur, 2016). MR images are also given importance in complete brain segmentation. Very few works reported brain parts segmentation(Ganesan and Radhakrishnan, 2009) using CT images. In some reported works skull removal and brain matter (BM) extraction from CT images is done as part of disease identification(Tang et al., 2011; Shirgaonkar et al. 2012) or segmentation(Cosic and Loucaric, 1997).

Some works reported extraction of BM from CT images using thresholding, region growing, midsagittal plane (MSP) finding techniques, supervised and unsupervised advanced mathematical approaches, shape guided segmentation and atlas-guided method. Most of the MSP finding techniques depend on brain symmetry(Brummer, 1991; Guillemaud et al., 1996; Davidson and Hugdahl, 1996; Ardekani et al., 1997) and thus not applicable for patients having visible brain abnormality. Less works are reported on segmentation of asymmetric brain using MSP finding. Segmentation of asymmetric brain using MSP finding technique suffers from higher error when midline shift and volume of abnormality is large. This method requires interpretation and larger time too(Hu and Nowinski, 2003).

Supervised techniques are basically classification based image segmentation in which training data are required to train the system for automatic segmentation. These train data are prepared by the expert by manual segmentation. Requirement of interpreting training data for every segmentation makes the process bulky. Low versatility in training dataset due to limited resources introduce errors in classification and thus in final segmentation too(Clarke et al., 1992). Paying the cost of complex algorithm, unsupervised segmentation techniques like k-mean segmentation (Kamble and Rathod, 2015), expectation maximization(Lee et al., 2008), fuzzy C means algorithms(Bezdek et al., 1993) are developed. For unsupervised method not training data set but parameter initialization is required. Shape guided segmentation(Neumann and Lorenz, 1998) or altas-guided methods are not much explored for CT BM extraction.

Thresholding and region growing are old, easy, simple and fast methods but not very potential for complex regions with overlapping or very close intensity regions. Thresholding can be done to segment entire image into two segments, one is above the threshold and another is below the threshold and the threshold value pixels can be combined with any of these two. A modification can be done by multithresholding(Sahoo et al., 1998) approach in which more than two segments can be extracted. Threshold can be user defined or automatically adapted from image (Shahangian and Pourghassem, 2015).

Region growing method is fast and offer higher accuracy for abnormalities with significant intensity difference from surroundings. Satisfactory results are reported in different CT segmentation works using region growing technique (Pohle and Toennies, 2001; Fresno et al., 2009; Wang et al., 2011). Region growing technique works based on pixel intensity or edge information. The major disadvantage with this method is the requirement of initial point or seed point definition by end user.

This paper has presented a comparatively robust and fast method of segmentation of BM from head CT scan. The proposed method can batch process multi-slice head CT of a patient in one run. Thresholding and region growing is used for mask creation and segmentation of 2D reference slice. Mask propagation is used to extract BM from other slices to reduce computational weight. Reference slice selection is done based on information of compactness of each slice. In this process clean extraction of BM is given most importance and thus any information outside the skull as well as the skull is removed completely. A modified region growing technique is used to make the process completely automatic. There is no need of Region of Interest (ROI) selection by user. Development of such a Computer Aided Diagnosis (CAD) system will be helpful for mankind as it will offer fast diagnosis with higher accuracy, lower subjectivity with no tiredness. Use of such system in hospitals with CT machine but no field expert or radiologist will offer initial diagnosis to fasten treatment. Different hospitals using same CAD will have homogeneous reports.

### 3. Methodology

The basic process flow is described in the flowchart given in figure1a. An entire dataset of multiple 2D scan is read, mask is formed, masking is done in two stages having an in-between step where the complete dataset is split into two parts with respect to a master slice selected automatically by CAD, at final stage holes are filled in masked images for information restoration. Detail steps of mask creation which includes thresholding, reference slices selection and computation of mask definition is shown in figure 1b.

#### 3.1 Database

Multiple slice head CT scan of different subjects is taken for segmentation of BM. Both diseased and normal scans are included in database. Total 28 patients' dataset, having more than 700 slices, are considered for this work. The number of slices varies from 23 to 34 for different subjects. The program is designed for complete automation; only the folder containing CT slices needs to be selected by the user. Extracted BM are named after their original name and saved automatically in a new folder in the same path of source folder. In this paper, the word 'dataset' presents all scanned CT slices of a patient in one sitting. The number of slices can vary depending on machine, practitioner's decision and machine capability. And the entire data used in this work is referred as 'database'.

#### 3.2 Selection of Reference Slice

For each dataset, dedicated master slice is selected from that dataset only. This master slice is used as reference for creation of mask definition. Use of global master or reference slice is avoided to make the system tolerant to the machine and/or scan specific differences in different dataset as slice from the same dataset under segmentation will have same artifacts related to machine like ring artifacts, noise, motion artifacts etc. as well as patient's alignment (Boas and Fleischmann, 2012). The complete dataset is read and thresholding is done by fixed threshold value 240 to remove the skull. The threshold value 240 is considered after examining large number of 8-bit grey scale brain CT images, having intensity distribution between 0 and 255, both normal and diseased. It has been observed that skull offers very high pixel intensity and no disease other than calcification offers intensity more than 240. This fixed threshold value is considered for skull removal as it shows high potential of clean removal of skull without affecting the brain, in large number of CT brain images from different sources. Calcifications, if any, are re-included later in the process of segmentation. Some image slices after applying threshold value are shown in figure2.

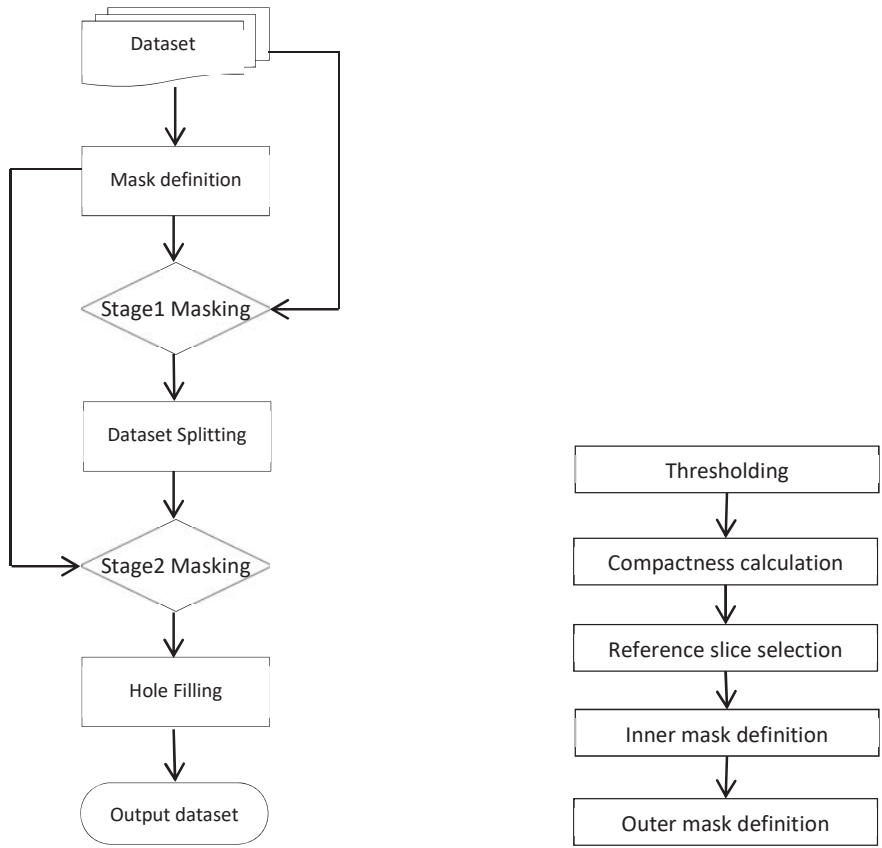

Figure 1 a: Process flowchart for brain segmentation from CT scan dataset  b: steps of mask definition creation

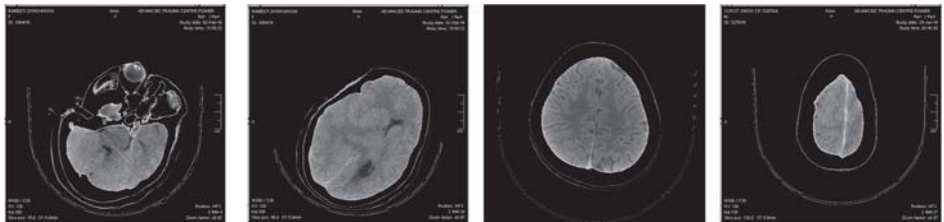

Figure 2: skull removed CT scan images

Not the original images but saved threshold images (TI) after skull removal are used for reference slice selection. These TIs majorly contain the brain matter along with lower intensity information outside the brain included during scan and processing of images. In a complete dataset facial parts are also included in some images where the brain area is low. The target is to select the CT slice having maximum brain area and BM information as reference. It has been observed that BM is the most compact part in any brain CT slice. In this paper, a simple but effective pixel level mathematical computation is introduced to identify most suitable reference slice.

For each TI of a dataset, compactness is evaluated. Here compactness is measured as the probability of each image pixel of having non-zero and non-background neighbor. It depends on the count of pixels having at least one adjacent pixel. Image having largest continuous area offers highest compactness. For an image $I(x,y)$ of size $n \times m$ compactness is, say, $C$, where $n$ is the number of rows and $m$ is the number of columns of the image array. $C$ can be described mathematically as follows –

$$C = \frac{1}{2(n*m)} * \left( \sum_{r=0}^{n} c_r + \sum_{c=0}^{m} c_c \right)$$

$$c_z = \sum_{i=0}^{u} a_i$$

$$a_i = \begin{cases} 1, & if\ p(i) \neq 0, p(i+1) \neq 0\ or\ p(i-1) \neq 0 \\ 0, & otherwise \end{cases}$$

$c_r$ and $c_c$ are total compactness count by row and column respectively.

$c_z = c_r$ for row wise computation and $c_c$ for column wise where $u = m$ for $c_r$ and $u = n$ for $c_c$

$a_i$ is the count of compactness of each series (row or column) of $I(x,y)$ of $u$ elements

$p(i)$ is $i^{th}$ element of $I(x,y)$

For simplification of computation another new method is introduced to handle index information of an array. A new 2D array is generated in which the array values are the corresponding index value of non-zero pixels of TI array. In the first step, a new array space is created using the dimension information of reference TI. In this array numeric value at each index is generated using index value of that particular index at which it is present. This array is addressed as index array (IA) in this paper. In the second step, the values at indices having 0 in reference TI are replaced by 0 in IA resulting a modified IA.

If the size of reference image I(x,y) is nXm, then the size of IA is also nXm. Use of this array, reduces the dimension of index of nXm 2D array from (n*m)X2 2D array to (n*m)X1 1D numeric value series and hence reduces the computational load and complexity. The adjacent values in row now are at unit distance. To find presence of neighbor a simple query runs row wise looking for difference 1 and column wise looking for difference $10^\ell$, where $\ell$ is dimension or length of largest column value. Formation of modified IA is discussed in steps below:

____________________________________

*Algorithm1 steps:*
1. Find $\ell$ where $\ell$ is the length of m i.e. $10^{(\ell -1)}$<m<$10^\ell$.
2. Form an array P of nXm size.
3. $P_{x,y}$= (x*$10^\ell$)+y; where x varies from 0 to n and y varies from 0 to m.
4. Put $P_{x,y}$=0 if $I_{x,y}$=0.
5. Run query row wise for neighbor i.e. p(i,j+1)-p(i,j)=1 or p(i,j)-p(i,j-1)=1
6. Run query column wise for neighbor i.e. p(i+1,j)-p(i,j)=$10^\ell$ or p (i,j)-p(i-1,j)=$10^\ell$
7. Sum-up all values, take average and divide by complete size of image i.e. n*m.

____________________________________

The 2D slice having highest compactness count from each dataset is selected as reference slice for segmentation of BM of that dataset. In this calculation, image texture or pixel intensity is not taken into account.

### 3.3 Creation of Mask

**3.3.1 Binary image creation**
The selected reference slice is processed to create mask for the complete dataset to segment BM from each 2D slice. The image converted into binary ($B_{ref}$) using the already used threshold for skull i.e. 240. All pixels having intensity below 240 are converted to 0 and others to 255 to get two distinct levels. Figure3 is clearly showing that the required BM is completely encircled by the white thick boundary represented by the skull of original image. Region growing technique with a seed point anywhere inside the brighten skull boundary can accurately define the region which can create mask for BM segmentation.

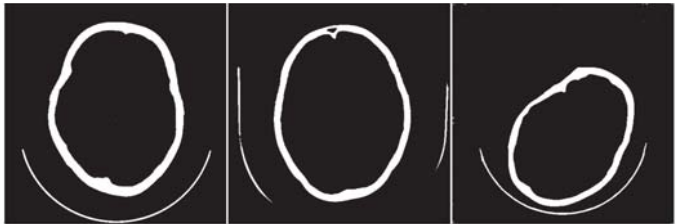

Figure 3: binary images of different reference slices

### 3.3.2 Seed point selection

As the reference slice contains large area of brain matter, non-zero pixels of that slice is mostly occupied by BM. To make the system completely automatic, instead of interactive seed point, a computational seed point finding method is used. Median of the image is found by separately locating median of row and median of column instead of locating intensity median. The intensity median can be somewhere outside of the BM area offering faulty segmentation. The proposed method will find the median nearest to the weight center of non-zero pixels. The TI of reference slice is used to locate median. For median computation, array index is important as only the presence of pixel not its value is counted. Hence the modified IA proposed above is considered for median finding. The method shows strong potential of placing seed point in the target is described below:

__________________________________________

*Algorithm2 steps:*
1. Access modified IA of previous step.
2. Sort the array.
3. Delete all 0s and get an array of indices of non-zero pixels only (0 majorly represents background).
4. Find median of this array. The value at median is representing the index of the median of reference TI.
5. The index is extracted by decomposition of value
   $n = $ quotient $(P_{x,y}/10^{\ell})$ and
   $m = $ remainder $(P_{x,y}/10^{\ell})$.
6. Select n as row median.
7. Repeat all 5 steps changing the dimension to mXn and select m from step 5 as column median.
8. Select (n,m) as seed point if the pixel value $P_{n,m} \neq 0$ or select nearest non-zero location as seed point.

__________________________________________

### 3.3.3 Mask Area Definition
Using the seed point information as origin and high pixel intensity i.e. 255 as stop criteria region growing technique is applied on $B_{ref}$ to segment the complete image into two regions – BM and the rest. Two different algorithms (algorithm 3 and algorithm4) are attempted for region growing. In the first algorithm, all similar intensity neighbors, in this case 0, are found and selected around the seed point. Then for each selected point the search repeats. The algorithm, described below, offers 100% accuracy in segmenting all pixels inside the brighten skull boundary at a cost of long execution time.

__________________________________________

*Algorithm3 steps*
1. Create an array G with an initial value $I_s$ which is index of $S_{x,y}$, the seed point.
2. Read value of all eight neighbors of $S_{x,y}$ i.e. of $P_{x+1,y}$, $P_{x-1,y}$, $P_{x+1,y+1}$, $P_{x+1,y-1}$, $P_{x-1,y+1}$, $P_{x-1,y-1}$, $P_{x,y+1}$, $P_{x,y-1}$.
3. Feed the index value of each location having similar intensity with $S_{x,y}$, i.e. 0, into G.
4. Read next index value (next to $I_s$ for the 1st time) in G and repeat step2 and 3.
5. Keep on repeating step 4 until inspection for the last element of G is done.
6. The final G contains index value of all pixels inside the brighten skull.

__________________________________________

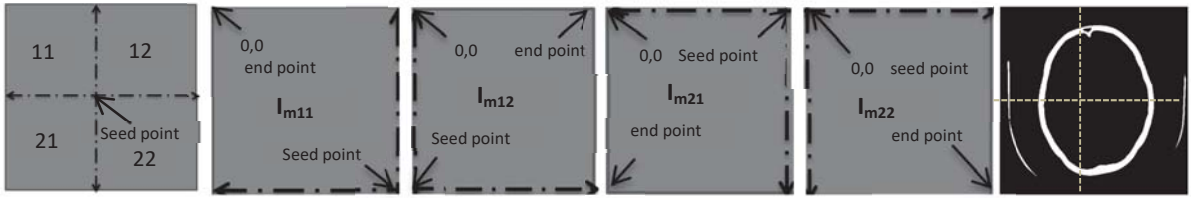

Figure 4: splitting an image into 4 sub-images and region growing seed pint for each sub-image

| Sub-image | Size | Seed point | Query end point |
|---|---|---|---|
| $I_{m11}$ | $n_1 \times m_1$ | $n_1-1, m_1-1$ | 0,0 |
| $I_{m12}$ | $n_2 \times m_2$ | $n_2-1, 0$ | $0, m_2-1$ |
| $I_{m21}$ | $n_3 \times m_3$ | $0, m_3-1$ | $n_3-1, 0$ |
| $I_{m22}$ | $n_4 \times m_4$ | 0,0 | $n_4-1, m_4-1$ |

Table 1: seed point of sub-images

As the execution time of this algorithm is significantly high, a second method is adopted. In this method taking seed point as origin, complete image is divided into four sub-images as shown in figure4. After region growing operation in each sub-image, these are merged back. This 'spilt and grow' algorithm works much faster than the above discussed neighbor searching algorithm.

Each sub-image is treated as an independent image by software during execution. So the indexing of each sub-image needs to be understood for proper handling and implementation of region growing technique. Four local seed points are required for four sub-images. For one sub-image it is the global seed point which serves as local; and for others, the corner nearest to the global seed point act as local seed point. For each sub-image the index value of that seed point is very different. And boundary search query direction also varies for each sub-image. The global seed point, extracted from algorithm2, may or may not be the image median; thus the size of sub-images may be equal or different. The local seed point and query end point for each sub-image is described in table1. Query runs in a single direction at a time. All columns are evaluated for each row and then rows are evaluated for each column of the image array. The mask generated by algorithm3 is considered as inner mask (IM).

___

*Algorithm4 steps*
1. Define split axes by seed point index.
2. Split image into 4 sub-regions by the axes.
3. Set seed point of each sub-image. It must be the global seed point or the corner point nearest to global seed point.
4. Identify all pixels of same intensity value within boundary i.e. enhanced skull. Query runs for simple value matching by row and column to keep the program fast and light weight.
5. Set a different fixed value, say1, at all identified location.
6. Marge sub-images back accurately with the help of split axes.
7. Remove hole, if any, within the grown region in the merged image.
8. Set all values other than the fixed value used to mark mask, here it is 1, to 0.

Output of step 8 is the mask for BM extraction of the dataset under segmentation.
___

IM works accurately for simple slices having BM part majorly, shown in figure5 (a) to (c). But slices having complex pattern due to inclusion of nasal area, sinuses, eye, optical canal etc. do not show good result due to presence of matching intensity of some non-BM parts as shown in figure5 (d) to (f). The BM shape also changes here fast in non-uniform pattern, first divided into lobs and then becomes very small in terminal scans. After IM masking, when a query runs to identify the adjacent areas outside mask, a large area is included. Extraction of actual BM is very difficult due to nonstandard pattern of shape change.

To reduce this error, concept of an additional mask, addressed as outer mask (UM) in this paper, is introduced. The outer mask is created using the IM of $B_{ref}$. The skull boundary part of reference IM is included in the mask area and converted into a larger mask. This wider area mask includes BM and thick skull border of the reference CT slice. This mask is used as global UM for dataset under segmentation. In case of UM, no progressive change is adapted.

The master UM is applied on each image slice of the dataset to remove any information outside the skull area of reference slice to avoid its inclusion during adjacent area query run. It reduces the area of non-BM part of complex nasal slices. Progressive UM can't work effectively as it keeps on including larger skull area which is not of our interest.

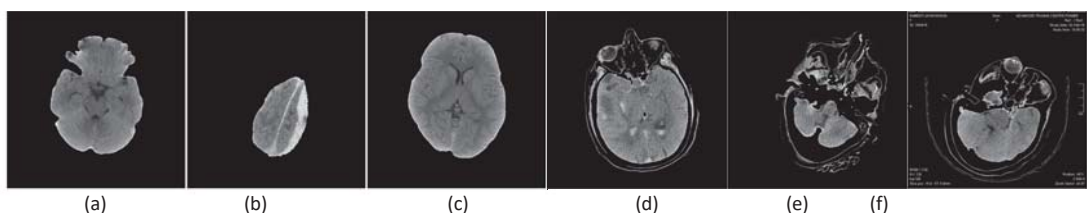

(a)　　　(b)　　　(c)　　　(d)　　　(e)　　　(f)

Figure 5: brain segmented by inner masking only

## 3.4 BM segmentation

After creation of IM and UM, all already saved TI slices of dataset are masked by UM. These outer masked images are then processed through IM; adjacent area query runs to extract BM part of each slice. In some terminal slices the BM remains small enough with respect to the definition of reference IM. Direct masking of those slices by IM collects unwanted information outside the skull. Few slices also have small extra BM area adjacent to defined IM. To avoid any error inclusion, accidental deletion of extra area and to get better segmentation, progressive IM modification is implemented. For a sequential scan from end to end the information pattern in the slices changes gradually from low to high to low. To take advantage of this pattern the complete dataset is divided into two parts one having slices from starting to reference slice another starting from reference to end. The slices just before and after the reference slice are executed by reference IM. After each masking, a query runs to check if any BM part of the segmented image is left outside but adjacent to the mask. Any such part left by masking is included. Part of the image, if any, under the definition of reference IM does not contain any BM information is excluded. A new IM is created from this modified definition. This modified mask is then used to segment next TI slice adjacent to it. In cases where the volume of modified mask is less than 5% of the reference IM the previous IM is used for next slice to avoid error due to non-overlapping small BM parts in scan. For dataset of 'n' number of images, if the reference IM slice is x, the propagation of mask will be from x to 0 and from x to n, where slices are ranked by the time of their creation. This time bound sorting helps to avoid wrong sequencing of slices, which cloud lead to faulty segmentation by progressive IM evolution.

Calcification information, if any, having high intensity within BM gets removed during thresholding because of its overlapping intensity range with skull. Such areas are represented by holes in the extracted BM. For accurate lossless segmentation it is required to remove each hole and include actual information. Using index information of hole(s) in extracted BM, the actual intensity value of calcification is collected from original image to replace the hole(s).

## 4. Result

Proposed algorithms are coded using LabVIEW (Laboratory Virtual Instrumentation Engineering Workbench) programming platform and implemented on 28 datasets of CT brain scan. In this complete collection some datasets are of normal subjects, rests are having some abnormality like calcification or hemorrhage. The proposed segmentation method is implemented on each dataset and the result of each step is discussed here following the sequence of program as shown in flowchart described in figure1.

### 4.1 Thresholding

Skull removed from each 2D image slice of complete dataset of each subject by removing all pixels above the proposed fixed threshold value 240. The result of different slices of different dataset is shown in figure6. It is seen that BM mostly remain unaffected by this thresholding. Only calcification is removed and in some hemorrhage cases small scattered holes are introduced in hemorrhage part due to its overlapping intensity components.

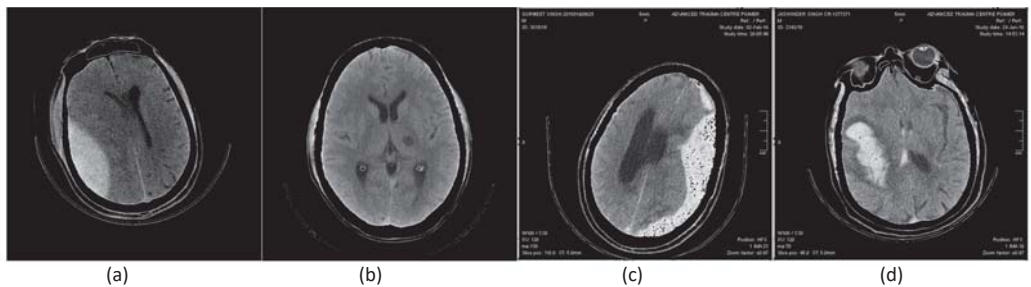

(a)  (b)  (c)  (d)

Figure 6: CT scan after thresholding

### 4.2 Reference slice selection

Form the threshold 2D slices reference slice is selected by computing compactness of each slice. In bare eye observation, it can be seen that the slices having simple pattern are more compact and the compactness count is higher for higher volume. Compactness value of two different complete dataset is shown in figure7 and figure8 along with the scan images. In each case one of the higher intracranial area slices is selected. Selected slices are highlighted by a boarder in the presented batches in figure7 and 8.

From the result it is clear that reference slice is selected irrespective of the location and sequence of the slice in the dataset. Though normally head CT scanning is done in sequence from one end to another end of head, the selection of middle slice in any dataset will not be wise as due to requirement of rescanning or some other reason this sequence can be violated. Selection of middle slice using the above stated concept offered selection of a null value slice from figure7 dataset which in turn offered no BM from the entire dataset, only because images in this dataset are not saved in the normally expected scanning sequence. Proposed compactness count offers an effective solution to this problem.

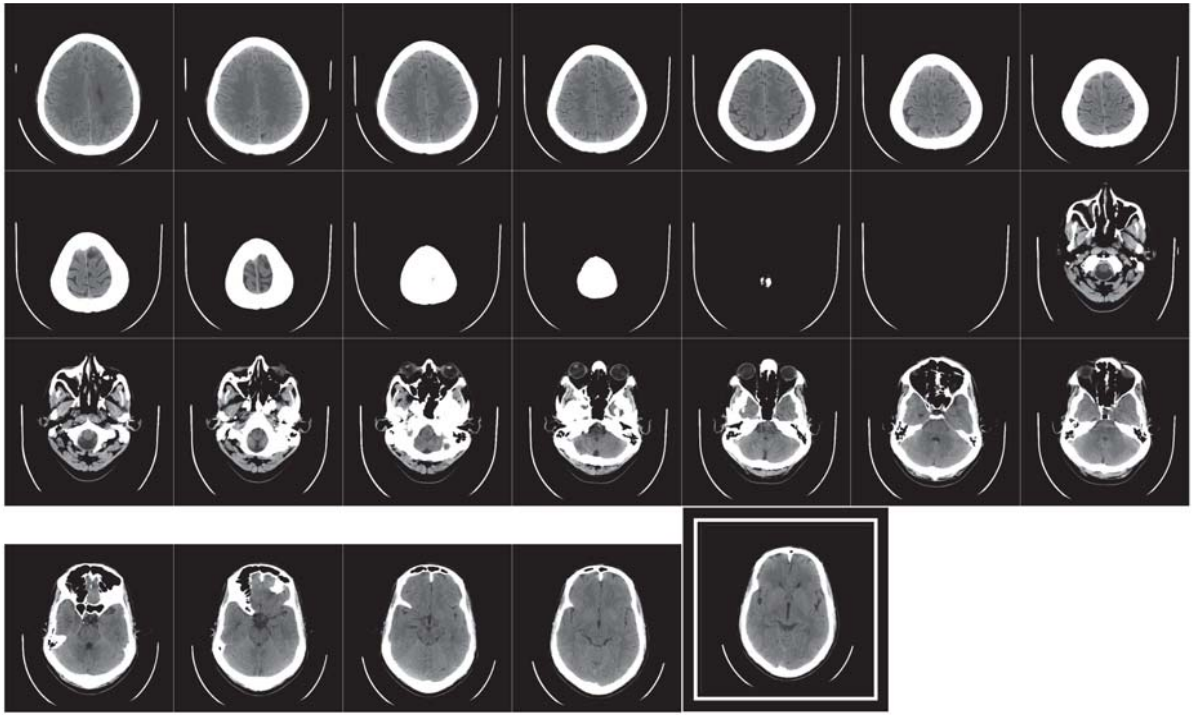
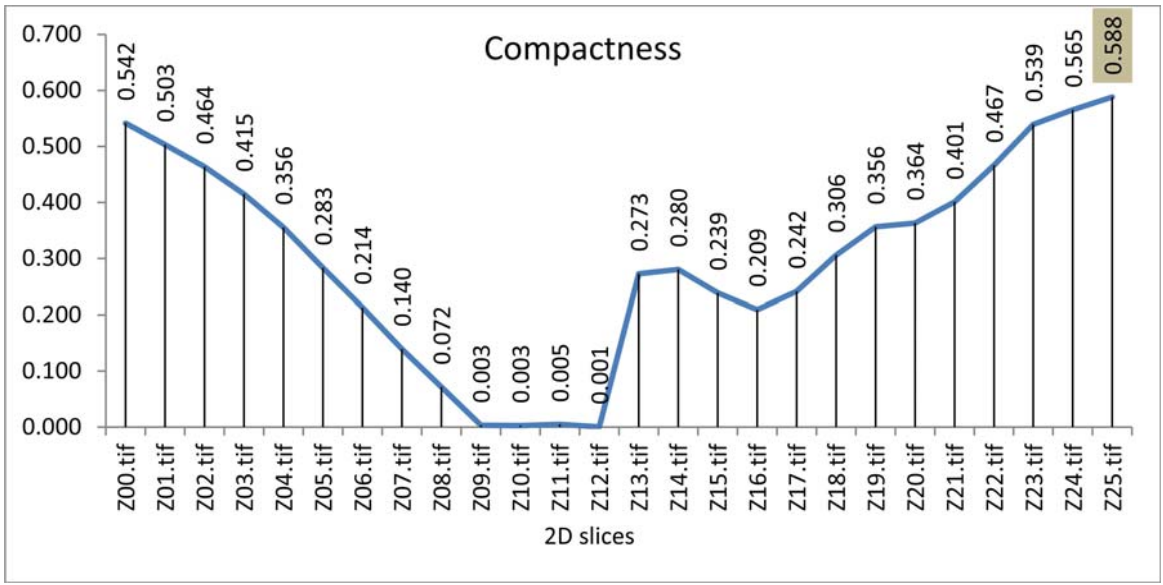

Figure 7: compactness value plot of given image dataset

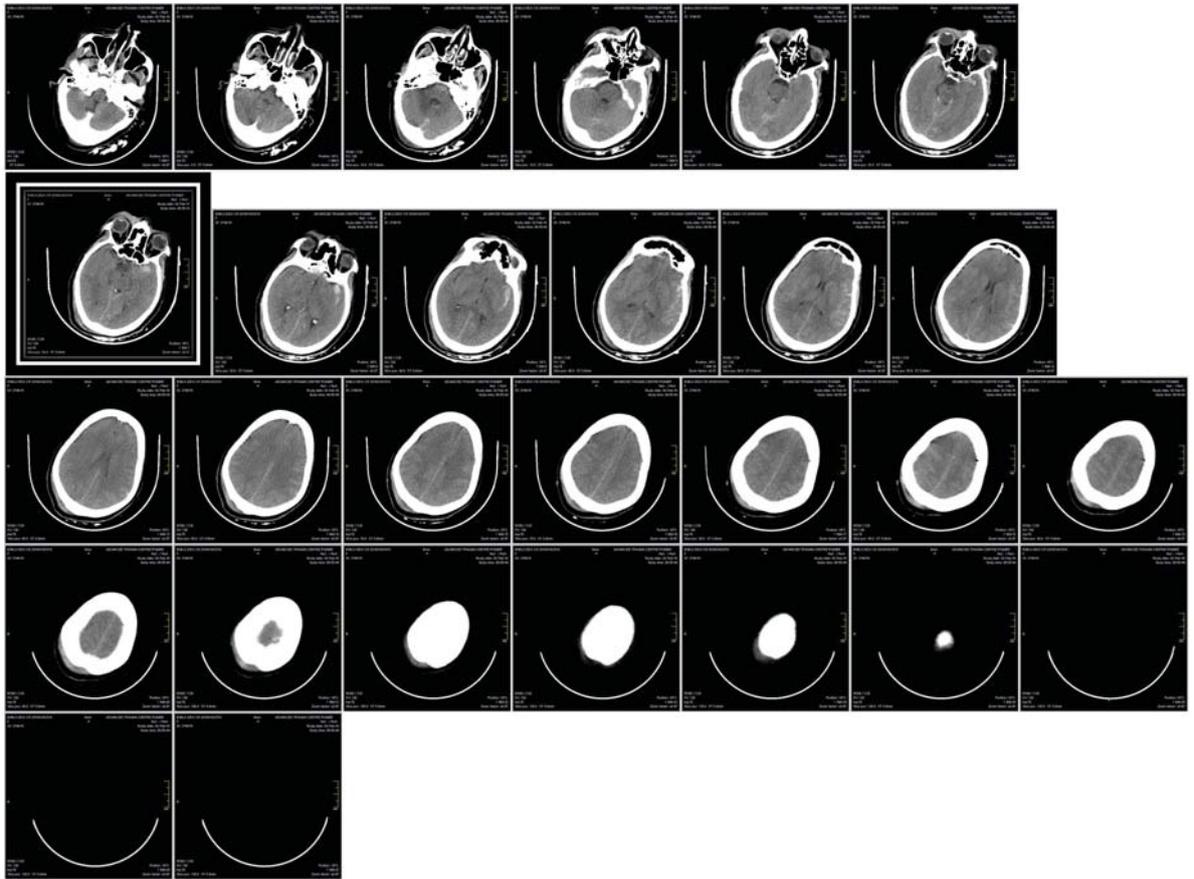
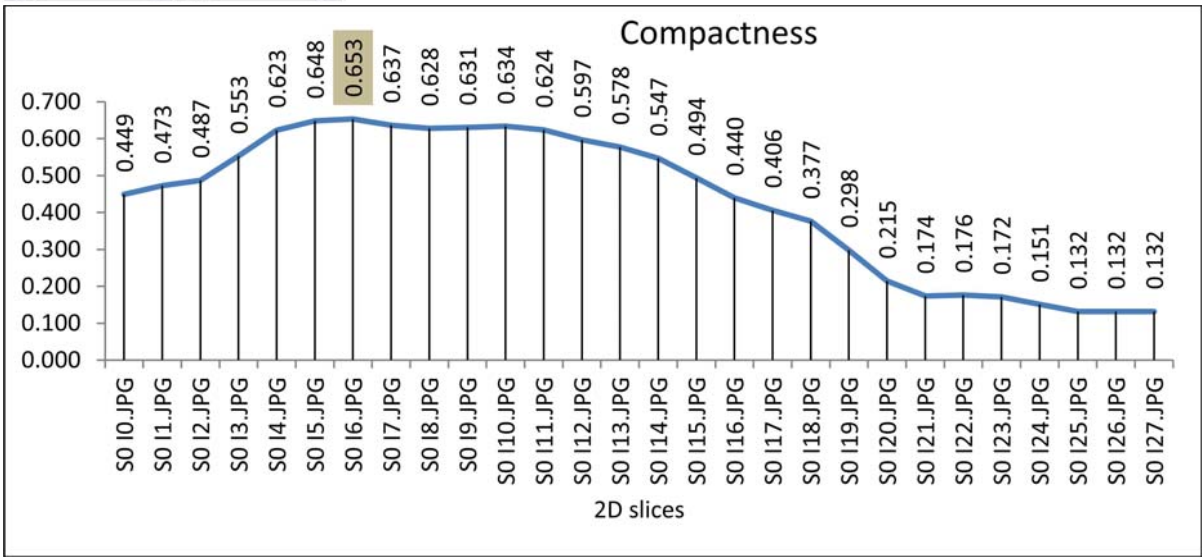

Figure 8: compactness value plot of given image dataset

### 4.3 Seed point & Mask

Center of splitting axes or the seed point is located by finding median of both axes of reference TI. In figure9 seed points are marked on reference images of the dataset shown above in figure7 and 8. In every case it is successfully placed within the intracranial location. Success rate is 100% for the 28 datasets under examination.

      Binary image formed from the reference slice by converting any value below threshold to zero and rest to 255. Then the binary image is split into four sub-images by the perpendicular axes originated from seed point (Δn,Δm). First the image is divided in two parts by x-axis at x=Δn i.e. one part contains 0 to Δn rows, another from (Δn+1) to n and then each part divided again by y-axis at Δm, splitting each from 0 to Δm and (Δm+1) to m when the complete image size is nXm. Region growing technique by pixel matching works accurately to define intracranial area for IM and skull included intracranial area for UM. Created IM and UM from the reference slices of figure9 are shown in figure10.

### 4.4 Segmentation

      Figure11 and figure12 present segmented BM of the respective dataset of masks presented in figure10 after masking and adjacent area search. Similar result for image slice with missing information (looked like small black holes) shown in figure6 (c) is also presented. The difference between before hole filling and after hole filling is visibly identifiable as shown in figure13.

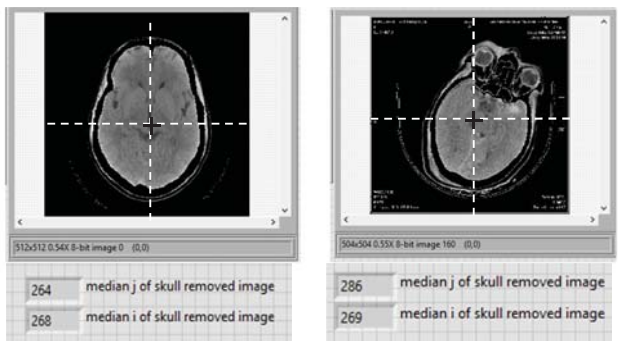

Figure 9: seed point location marked by ✛

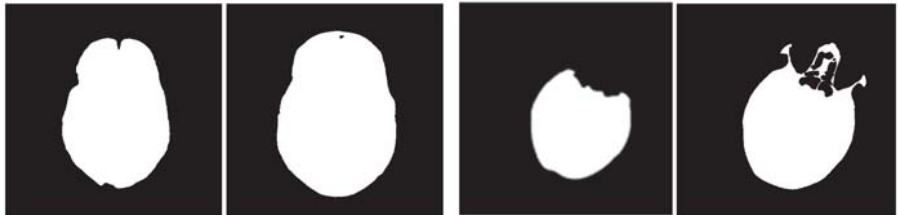

Figure 10: IM and UM of fig.7 and fig.8 dataset respectively

      Complete segmented dataset presents the method's performance quality. All slices, having brain parts majorly, offer very clean segmentation. Slices beyond any null BM slice are also segmented accurately. Some nasal slices are offering larger area than actual BM area. Lower intensity value muscle areas around skull and facial bones get included. Segmentation by progressive mask does not include bone adjacent muscles; but during left out adjacent search, these areas get included due to its connectivity with BM area in the scan.

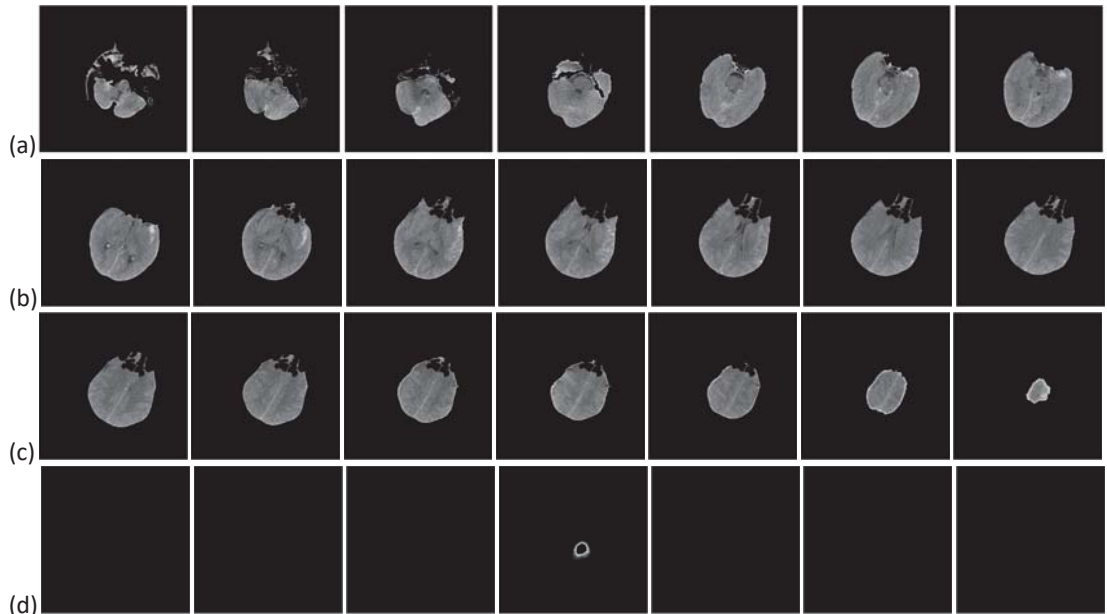
Figure 11: segmented brain for fig.7 dataset

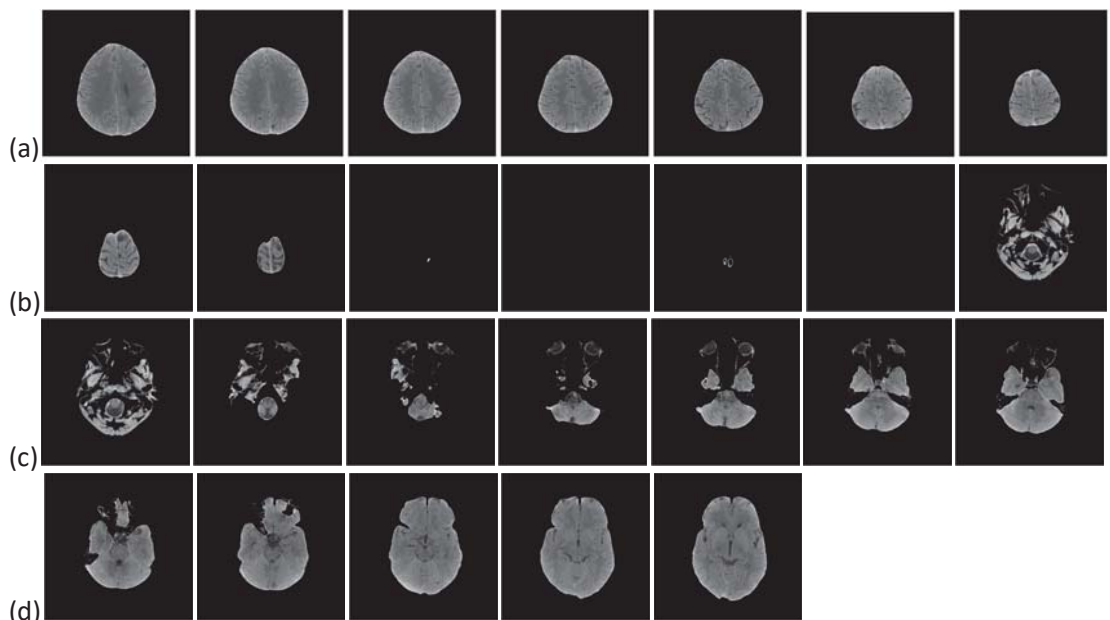
Figure 12: segmented brain for fig.8 dataset

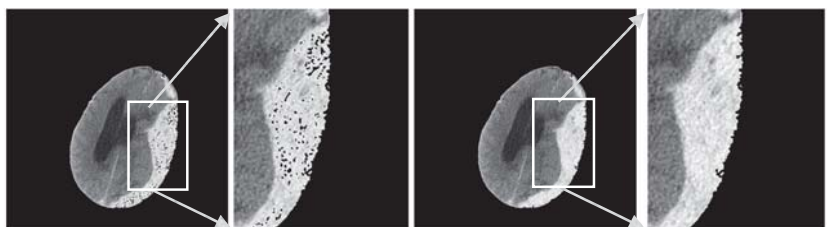
Figure13: hole filling at final stage

## 4.5 Result analysis

Proposed methods performance analysis is done in terms of accuracy, sensitivity, specificity, error, PPV (positive predictive value), NPV (negative predictive value) using the values of true positive (TP), false positive (FP), false negative (FN), true negative (TN). TP and TN present correct segmentation (CS) as per clinical ground truth. TN is the count of slices in segmented dataset which contain no BM because of having no intracranial information in corresponding original slice. The inclusion of extra tissue parts in nasal area and having non-zero segmented values for original slices having no intracranial information is addressed as FP. In FP, segmented slices having extra tissue parts in nasal area also contribute in TP count as there are BM parts too. Here thus another term absolute FP (AbsFP) is calculated to identify slices having non-zero value without BM. The conceptual presentation of TP, FP, TN and FN is presented pictorially in figure14 where FP basically presents AbsFP. The overall segmentation performance is calculated with respect to the volume of a dataset i.e. the total number of slices. For a dataset of n number of slices the analysis parameter are extracted as follows:

$$CS = TP + TN$$

$$n = AbsFP + FN + CS$$

$$\%error = \frac{AbsFP + FN}{n} * 100\%$$

$$\%accuracy = \frac{CS}{n} * 100\%$$

$$sensitivity = \frac{TP}{TP + FN}$$

$$specificity = \frac{TN}{TN + AbsFP}$$

$$positive\ predictive\ value = \frac{TP}{TP + FP}$$

$$negative\ predictive\ value = \frac{TN}{TN + FN}$$

The proposed method has offered zero FN segmentation i.e. the sensitivity turns into 1 for this method. For the dataset having no slice without BM TN turns into zero. The specificity and NPV are not calculated for such dataset. BM segmentation using proposed method demonstrated very high accuracy. The complete analysis result is presented in table2. Serial no. 24 showing very poor performance due to faulty dataset presented in figure15. Dataset15 is showing comparatively lower accuracy due to higher AbsFP value contributed by soft tissues in no-BM slices.

|  | Slice with BM | Slice without BM |
|---|---|---|
| Segmentation have BM | TP | FP |
| Segmentation without BM | FN | TN |

Figure 14: pictorial presentation of TP, FP, FN, TN

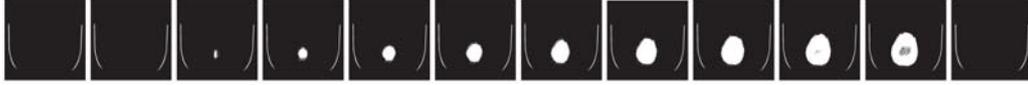
Figure 15: faulty dataset

| dataset serial no. | TP | FN | AbsFP | TN | N | CS | sensitivity | specificity | PPV | NPV | %error | % accuracy |
|---|---|---|---|---|---|---|---|---|---|---|---|---|
| 1 | 22 | 0 | 0 | 2 | 24 | 24 | 1 | 1 | 1 | 1 | 0 | 100 |
| 2 | 28 | 0 | 1 | 5 | 34 | 33 | 1 | 0.833 | 0.966 | 1 | 2.941 | 97.059 |
| 3 | 28 | 0 | 0 | 0 | 28 | 28 | 1 | -- | 1 | -- | 0 | 100 |
| 4 | 21 | 0 | 1 | 6 | 28 | 27 | 1 | 0.857 | 0.955 | 1 | 3.571 | 96.429 |
| 5 | 32 | 0 | 0 | 0 | 32 | 32 | 1 | -- | 1 | -- | 0 | 100 |
| 6 | 25 | 0 | 0 | 0 | 25 | 25 | 1 | -- | 1 | -- | 0 | 100 |
| 7 | 23 | 0 | 0 | 3 | 26 | 26 | 1 | 1 | 1 | 1 | 0 | 100 |
| 8 | 21 | 0 | 1 | 4 | 26 | 25 | 1 | 0.8 | 0.955 | 1 | 3.846 | 96.154 |
| 9 | 19 | 0 | 1 | 6 | 26 | 25 | 1 | 0.857 | 0.95 | 1 | 3.846 | 96.154 |
| 10 | 24 | 0 | 0 | 2 | 26 | 26 | 1 | 1 | 1 | 1 | 0 | 100 |
| 11 | 23 | 0 | 1 | 2 | 26 | 25 | 1 | 0.667 | 0.958 | 1 | 3.846 | 96.154 |
| 12 | 21 | 0 | 1 | 3 | 25 | 24 | 1 | 0.75 | 0.955 | 1 | 4 | 96 |
| 13 | 25 | 0 | 0 | 0 | 25 | 25 | 1 | -- | 1 | -- | 0 | 100 |
| 14 | 24 | 0 | 1 | 1 | 26 | 25 | 1 | 0.5 | 0.96 | 1 | 3.846 | 96.154 |
| 15 | 24 | 0 | 2 | 0 | 26 | 24 | 1 | -- | 0.923 | -- | 7.692 | 92.308 |
| 16 | 24 | 0 | 1 | 1 | 26 | 25 | 1 | 0.5 | 0.96 | 1 | 3.846 | 96.154 |
| 17 | 23 | 0 | 1 | 2 | 26 | 25 | 1 | 0.667 | 0.958 | 1 | 3.846 | 96.154 |
| 18 | 19 | 0 | 1 | 3 | 23 | 22 | 1 | 0.75 | 0.95 | 1 | 4.348 | 95.652 |
| 19 | 24 | 0 | 0 | 0 | 24 | 24 | 1 | -- | 1 | -- | 0 | 100 |
| 20 | 23 | 0 | 0 | 3 | 26 | 26 | 1 | 1 | 1 | 1 | 0 | 100 |
| 21 | 26 | 0 | 0 | 0 | 26 | 26 | 1 | -- | 1 | -- | 0 | 100 |
| 22 | 26 | 0 | 0 | 0 | 26 | 26 | 1 | -- | 1 | -- | 0 | 100 |
| 23 | 26 | 0 | 0 | 0 | 26 | 26 | 1 | -- | 1 | -- | 0 | 100 |
| 24 | 2 | 0 | 3 | 7 | 12 | 9 | 1 | 0.7 | 0.4 | 1 | 25 | **75** |
| 25 | 23 | 0 | 0 | 0 | 23 | 23 | 1 | -- | 1 | -- | 0 | 100 |
| 26 | 26 | 0 | 0 | 0 | 26 | 26 | 1 | -- | 1 | -- | 0 | 100 |
| 27 | 26 | 0 | 0 | 0 | 26 | 26 | 1 | -- | 1 | -- | 0 | 100 |
| 28 | 26 | 0 | 0 | 0 | 26 | 26 | 1 | -- | 1 | -- | 0 | 100 |

Table 2: analysis of segmentation performance of proposed method

## 5. Discussion and Conclusion

The proposed method is fast, reliable and robust. It can segment any given dataset automatically with no false negative. To enhance the speed, two dimensional Indices of 2D array are converted into single numerical values suitable for fast and simple mathematical computation; an automatic seed point selection is proposed to take advantages of fast and simple region grow technique. The compactness count offers automatic selection of most suitable slice in any dataset for mask definition creation. Proposed continuous propagation and modification method of inner mask, results into an adaptive masking of higher efficiency. The outer mask helps to eliminate headrest, embedded information and some peripheral parts from nasal slices.

During speed test, a complete dataset of 34 slices got segmented in less than 11 seconds in a computer having 64-bit operating system, 8 GB RAM and processor of Intel® Core™ i7-3770 CPU @ 3.40GHz. Because of having guaranteed '0' FN, there is 0% chance of missing BM in any slice. During segmentation enhancement, averaging, noise reduction operations which can modify pixel value of an image is avoided intentionally to keep the segmented output unaltered with respect to dataset images offering 0% chance of removal of or change in diagnostic information.

The segmentation result of proposed method supports disease identification of any kind of brain disease visible in brain CT as the complete BM part segmented successfully. The only issue is inclusion of non-intracranial information in few nasal slices. This will have less effect during disease detection, if disease location identified on another slice and then a query based on disease location and texture information of that slice runs through adjacent slices. It will narrow down search area and will not be much affected by inclusion of non-intracranial parts. Till, a scope of further work is left for better segmentation of these slices to identify BM precisely.

## Acknowledgement

This research did not receive any specific grant from funding agencies in the public, commercial, or not-for-profit sectors. The image data are provided by Department of Radiodiagnosis, PGIMER (Postgraduate Institute of Medical Education and Research) Chandigarh, India and Dr. Tanmay Ghosh of Nirnoy CT Scan Center, Bishnupur, West Bengal, India. Support in image interpretation and image evaluation is provided by Department of Radiodiagnosis, PGIMER.

## References

Al-Ayyoub, M., Alawad, D. U. A. A., Al-Darabsah, K., and Aljarrah, I. N. A. D., 2013. Automatic detection and classification of brain hemorrhages. WSEAS Transactions on Computers. 12(10), 395-405.

Anand, Ashima, and Harpreet Kaur, 2016. Survey on Segmentation of Brain Tumor: A Review of Literature. International Journal of Advanced Research in Computer and Communication Engineering. Vol. 5(1), 79-82.

B.A. Ardekani, J. Kershaw, M. Braun, I. Kanno, 1997. Automatic detection of mid-sagittal plane in 3D brain images. IEEE Transact. Med. Imag., 16 (6), 947–952.

Bardera, A., Boada, I., Feixas, M., Remollo, S., Blasco, G., Silva, Y. and Pedraza, S., 2009. Semi-automated method for brain hematoma and edema quantification using computed tomography. Computerized Medical Imaging and Graphics. 33(4), 304-311.

Boas, F.E. and Fleischmann, D., 2012. CT artifacts: causes and reduction techniques. Imaging in Medicine. 4(2), 229-240.

Brummer, M.E., 1991. Hough transform detection of the longitudinal fissure in tomographic head images. IEEE Transact. Med. Imag. 10 (1),4–81.

Chen, W., Smith, R., Ji, S.Y., Ward, K.R. and Najarian, K., 2009. Automated ventricular systems segmentation in brain CT images by combining low-level segmentation and high-level template matching. BMC medical informatics and decision making. 9(1), 1.

Clarke, L.P., Velthuizen, R.P., Hall, L.O., Bezdek, J.C., Bensaid, A.M. and Silbiger, M.L., 1992. Comparison of supervised pattern recognition techniques and unsupervised methods for MRI segmentation. In Medical Imaging. International Society for Optics and Photonics. 6, 668-677.

Cosic, D. and Loucaric, S., 1997. Computer system for quantitative: analysis of ICH from CT head images. Engineering in Medicine and Biology Society. Proceedings of the 19th Annual International Conference of the IEEE. 2, 553-556.

Davidson, R.J., Hugdahl, K., 1996. Brain Asymmetry. MIT Press/Bradford Books, Cambridge, MA.

del Fresno, M., Vénere, M. and Clausse, A., 2009. A combined region growing and deformable model method for extraction of closed surfaces in 3D CT and MRI scans. Computerized Medical Imaging and Graphics. 33(5), 369-376.

Ganesan, R. and Radhakrishnan, S., 2009. Segmentation of computed tomography brain images using genetic algorithm. International Journal of Soft Computing. 4(4), 157-161.

Guillemaud, R., Marais, P., Zisserman, A., McDonald, B., Crow, T.J., Brady, M., 1996. A three dimensional mid sagittal plane for brain asymmetry measurement. Schizophr. Res. 18 (2-3), 183–184.

Hu, Q. and Nowinski, W.L., 2003. A rapid algorithm for robust and automatic extraction of the midsagittal plane of the human cerebrum from neuroimages based on local symmetry and outlier removal. NeuroImage. 20(4), 2153-2165.

Huang, Y. and Parra, L.C., 2015. Fully automated whole-head segmentation with improved smoothness and continuity, with theory reviewed. PloS one. 10(5), p.e0125477.

J.C. Bezdek, L.O. Hall, L.P. Clarke, 1993. Review of MR image segmentation techniques using pattern recognition. Med. Phys. 20, 1033–1048.

Kamble, Sanghamitra T., and M. R. Rathod., 2015. Brain Tumor Segmentation using K-Means Clustering Algorithm. International Journal of Current Engineering and Technology. 5(3), 1521-1524.

Lee, Tong Hau, Mohd FA Fauzi, and Ryoichi Komiya, 2008. Segmentation of CT brain images using K-means and EM clustering. Computer Graphics, Imaging and Visualisation. Fifth International Conference. CGIV'08, 339-344.

Neumann A, Lorenz C. 1998. Statistical shape model based segmentation of medical images. Comput. Med. Image Graph. 22,133–43

Pohle, R. and Toennies, K.D., 2001. Segmentation of medical images using adaptive region growing. In Medical Imaging. International Society for Optics and Photonics. 1337-1346.

Rehana, K., Tabish, S., Gojwari, T., Ahmad, R. and Abdul, H., 2013. Unit cost of CT scan and MRI at a large tertiary care teaching hospital in North India. Health. 5, 2059-2063

Sahoo, Prasanna K., S. A. K. C. Soltani, and Andrew KC Wong, 1988. A survey of thresholding techniques. Computer vision, graphics, and image processing. 41.2, 233-260.

Shahangian, B. and Pourghassem, H., 2016. Automatic brain hemorrhage segmentation and classification algorithm based on weighted grayscale histogram feature in a hierarchical classification structure. Biocybernetics and Biomedical Engineering. 36(1),217-232.

Shirgaonkar, S., Jeong, D.H., Huynh, T. and Ji, S.Y., 2012. Designing a robust bleeding detection method for brain CT image analysis. Bioinformatics and Biomedicine Workshops (BIBMW). IEEE International Conference. 260-264.

Tang, Fuk-hay, Douglas KS Ng, and Daniel HK Chow, 2011. An image feature approach for computer-aided detection of ischemic stroke. Computers in biology and medicine. 41.7, 529-536.

Varma DR., 2012. Managing DICOM images: Tips and tricks for the radiologist. The Indian Journal of Radiology & Imaging. 22(1), 4-13.

Wang, H.R., Yang, J.L., Sun, H.J., Chen, D. and Liu, X.L., 2011. An improved region growing method for medical image selection and evaluation based on Canny edge detection. In Management and Service Science (MASS), International Conference. 1-4.